\documentclass[runningheads]{llncs}
\usepackage[T1]{fontenc}
\usepackage{graphicx}
\usepackage{booktabs} 
\usepackage{multirow}
\usepackage{diagbox}
\usepackage{subfig}
\usepackage{amsmath}
\usepackage{xcolor}

\begin{document}
\title{Exploring The Neural Burden In Pruned Models: An Insight Inspired By Neuroscience}
\titlerunning{Neural Burden}
%
\author{Zeyu Wang\inst{1}$^{*}$ \and
Weichen Dai\inst{1}\thanks{Equal contributions.} \and
Xiangyu Zhou\inst{2} \and
Ji Qi\inst{2}$^{**}$ \and
Yi Zhou\inst{1}\thanks{Corresponding authors.}}
\authorrunning{Wang et al.}
%
\institute{University of Science and Technology of China \and
China Mobile(SuZhou)Software Technology Co.,Ltd. \\
\email{\{wangzy1822,dddwc\}@mail.ustc.edu.cn \\ 
\{zhouxiangyu,qiji\}@cmss.chinamobile.com \\
\{yi\_zhou\}@ustc.edu.cn }}
\maketitle              

\begin{abstract}
Vision Transformer and its variants have been adopted in many visual tasks due to their powerful capabilities, which also bring significant challenges in computation and storage.
Consequently, researchers have introduced various compression methods in recent years, among which the pruning techniques are widely used to remove a significant fraction of the network. 
Therefore, these methods can reduce significant percent of the FLOPs, but often lead to a decrease in model performance.
To investigate the underlying causes, we focus on the pruning methods specifically belonging to the pruning-during-training category, then drew inspiration from neuroscience and propose a new concept for artificial neural network models named \textbf{Neural Burden}.
We investigate its impact in the model pruning process, and subsequently explore a simple yet effective approach to mitigate the decline in model performance, which can be applied to any pruning-during-training technique.
Extensive experiments indicate that the neural burden phenomenon indeed exists, and show the potential of our method.
We hope that our findings can provide valuable insights for future research.
Code will be made publicly available after this paper is published.

\end{abstract}

\section{Introduction}
Recently, Vision Transformer (ViT)\cite{dosovitskiy2021image} and its variants\cite{Dong_2022_CVPR,Li_2022_CVPR,touvron2021training,yuan2022volo} have shown impressive performance in various computer vision tasks like image classification, object detection, and image segmentation. 
As the field of large models continues to evolve, ViT is becoming a popular choice as the foundational visual model for large visual and multimodal models\cite{alayrac2022flamingo,kirillov2023segany,girdhar2023imagebind}.
However, the significant computational and storage requirements pose a pressing challenge\cite{chen2021chasing,kong2022spvit} that must be addressed for efficient training and deployment of these models.

Tremendous advances, including pruning, parameter sharing (quantization), and low-rank factorization, have been made recently, among which the pruning techniques are widely utilized for a lightweight model.
Most of the early pruning methods for Transformer-architecture models drew on the rich experience of compressing convolutional neural networks.
These methods obtained compressed neural networks by using binary masks or direct pruning to reduce model parameters and artificial neuron connections\cite{zhu2021vision,wang2022vtc,zheng2022savit,yu2022unified}. 
Considering the structural characteristics of transformer models and their tokenized data, recent work\cite{rao2021dynamicvit,xu2022evo,li2023constraint,bolya2022token} has increasingly focused on data compression techniques to achieve model acceleration and compression. 
For example, within ViTs, not all tokens contribute equally to the final task, and there is often a high degree of similarity among some tokens. 
Therefore, computational cost can be reduced by deleting or merging tokens.
Moreover, some researchers have attempted to combine structural compression with token pruning, which achieve promising results\cite{hou2022multi,wang2022vtc,wang2023cait}.
Although the aforementioned methods have shown comparable performance in reducing computational cost and storage cost, they usually experience a decline in performance most of the time, especially in highly sparse scenarios.

Limited research has been conducted to develop a theoretical understanding of the impact of model compression. 
\cite{gao2019rate} offers an information-theoretic perspective on model compression through rate-distortion theory, emphasizing the tradeoff between model compression and the empirical risk of the compressed model. 
While in \cite{dziugaite2017computing,zhou2018non}, a PAC-Bayesian framework is employed to derive a non-vacuous generalization error bound for the compressed model, attributing this to its reduced model complexity.
\cite{bu2021population} use the mutual information based generalization error bound jointly with rate distortion theory to connect analyses of generalization error and empirical risk.

On the other hand, neuroscience research aims to uncover the reasons behind the remarkable power of biological neurons and nervous systems, which is naturally valuable on providing inspiration for deep learning studies. 
Numerous works have leveraged advancements in neuroscience to propose more efficient and effective neural network models or algorithms.
For example, based on the observation that the brain often solves decision-making with an evidence accumulation mechanism, DDDM\cite{chen2022dddm} integrates test-phase dropout with the Dynamic Dropout Mechanism (DDM) to enhance the robustness of arbitrary neural networks.
Therefore, we aim to delve into neuroscientific research to find possible explanations for the weakened performance caused by model pruning.
Inspired by this motivation, we propose a new concept on neural network models named \textbf{Neural Burden}, which means that during the pruning process, artificial neurons in network are burdened because of the need to compensate for the lost information and the need of adaptation to the given data, leading to a weakened performance.
We examine our conjecture and further design a simple yet effective method to help mitigate the degradation.
Specifically, we applied our approach to a four-layer ViT model and test it on the CIFAR-10 dataset.
The experimental results indicate the potential of our proposed method.

\section{Related works}
\paragraph{\textbf{Vision Transformer.}}
Before introducing Transformers\cite{vaswani2017attention} into the field of computer vision (CV), various models based on the Transformer architecture had already achieved significant success in natural language processing (NLP). 
By stacking models and utilizing large amounts of corpus data for pretraining, these models\cite{devlin2018bert} with large numbers of parameters could achieve outstanding performance after fine-tuning on relatively small-scale task-specific datasets.
Adopting a similar approach, Vision Transformers (ViTs)\cite{dosovitskiy2021image} have rapidly achieved performance in the field of computer vision (CV) that rivals or even surpasses that of convolutional neural networks (CNNs). 
ViTs have demonstrated exceptional performance in tasks such as image classification\cite{chen2021crossvit,chen2022fast}, object detection\cite{horvath2021manipulation,wang2022bridged}, image segmentation\cite{hatamizadeh2022unetformer,yang2022lavt}, image reconstruction\cite{chen2021pre,yang2020learning}, and 3D point cloud processing\cite{wang2022bridged}. 
Subsequently, a series of variants\cite{Dong_2022_CVPR,Li_2022_CVPR,touvron2021training,yuan2022volo} based on the ViT architecture have further extended the influence of the transformer architecture. 
In many of the current multimodal large models\cite{alayrac2022flamingo,kirillov2023segany,girdhar2023imagebind}, ViTs often serve as the backbone model for the visual modality, playing a crucial role.
However, as the performance of ViTs continues to improve, the dense computational cost and high storage requirements have become challenging issues. 
This has driven ongoing research into lightweight and sparse ViT models, leading to the development of a series of effective methods\cite{zhu2021vision,zheng2022savit,yu2022unified,bolya2022token}.

\paragraph{\textbf{Model Pruning.}}
Pruning is a kind of compression methods used for almost all models, aiming to remove redundant connections between artificial neurons to achieve model sparsity. 
There are three main categories of pruning strategies, namely pruning before training, pruning during training and pruning after training.
Thanks to the rich experience in compressing CNN models accumulated in the computer vision field, many pruning methods have been transferred to ViT models for compression.
Existing pruning methods typically use certain evaluation metrics to measure the importance score of each parameter, and then decide to retain or remove the corresponding weights based on these scores, resulting in a sparse network. 
For example, importance can be assessed by the magnitude of the weights or gradients\cite{lee2018snip,tanaka2020pruning,Lin2020Dynamic}. 
Similar to the establishment and dynamic learning process of brain synapses, many algorithms attempt to implement pruning during training (PDT).
Some methods dynamically adjust the parameter mask during the iteration, while others use learnable Gumbel Softmax to achieve automatic mask learning.

\paragraph{\textbf{Neuroscience}}
The brain, as a complex network system, exhibits sparse connectivity between neurons\cite{foldiak2003sparse}. 
During the learning and remodeling processes of biological neural networks, some neurons take over the functions of those that are removed, thereby maintaining overall network stability and functionality. 
For example, in the vertebrate nervous system, Mautner cells' function can still be preserved when these cells are partially removed\cite{doi:10.1073/pnas.1918578117}.
This supports the notion that the brain can self-compensate by adjusting the balance of excitation and inhibition within neural circuits. 
This compensation mechanism relies on the reconfiguration among existing neurons, and when this balance is disrupted, the compensatory ability is limited, potentially leading to functional deficits and the emergence of symptoms\cite{barrett2016optimal}.

Due to the limited bandwidth of information transmission, the brain must compress and optimize information processing\cite{troscianko2023model}. 
The load theory of attention suggests that as perceptual load increases, attention focuses on the most relevant information, ignoring distractors. 
Under high perceptual load conditions, information processing becomes more focused and efficient\cite{Blindedbytheload}. 
This theory helps explain how the brain maintains cognitive performance by compressing information under varying levels of perceptual load. 
Similarly, the load of visual working memory significantly affects the processing of other sensory information. 
For instance, when the visual working memory load increases, the ability to process auditory stimuli may decrease\cite{effectsofvisualworkingmemoryload}. 
This indicates that during reasoning and cognition, the brain maintains network stability by adjusting its information processing strategies.

These findings demonstrate that during the learning and reasoning processes of biological neural networks, the brain maintains overall network functionality and stability by regulating the balance between neuron load and input information. 
These insights can inspire research into exploring the performance degradation observed in artificial neural networks during compression, and motivate research into improving existing compression algorithms to reduce the losses incurred during this process.

\section{Methodology}

We first attempt to find similar phenomena and possible explanations from neuroscience. 
Subsequently, we seek to draw on studies of neuronal characteristics in neuroscience to propose a simple yet effective enhancement method for pruning.
Our discussion focuses on the pruning-during-training methods in this paper, and we leave the other settings for future work.

\subsection{Investigating the Neural Burden}
A natural idea for the reasons of performance degradation caused by model compression is the reduction in the number of parameters.
However, compressed networks have been observed to exhibit superior generalization compared to full models in some cases\cite{mocanu2018scalable}.
Moreover, studies in neuroscience have shown that brain has an impressive ability to withstand neural damage up to a certain point \cite{barrett2016optimal}.
These two consistent observations inspired us to seek other potential explanations for the decline in model performance from established conclusions in neuroscience.
As described in related work, the theory of perceptual load in neuroscience indicates that biological neurons possess a limited bandwidth and may loss information under high-load conditions.
Therefore, we argue that the main reason for declined performance lies in the limited conductive capacity of artificial neurons, which means during the pruning process, neurons in artificial network need to self-regulate to compensate for the loss of information representation while further fitting the given training data, resulting in heavy burden to the artificial neurons.

To prove our hypothesis, we first introduce a cost-benefit trade-off of neural system representation performance in neuroscience from \cite{barrett2016optimal}:
\begin{equation}
    E = (s - \hat{s})^2 + \beta * C(r),
\end{equation}
where $s$ is the actual signals, $\hat{s}$ for the readout, $\beta$ stands for the trade-off, and $C(r)$ quantifies the cost of the representation with $r$ being neurons’ instantaneous firing rates.
We believe that the deep learning methods intrinsically have a similar dynamic.
From the perspective of artificial neural network, the first term quantifies the preserved information, and the second term quantifies sparseness of network parameters which is beneficial since sparsity in many forms may be helpful with robustness\cite{ahmad2019can}.
Thus, we borrow the idea from \cite{olshausen1996emergence} to further evaluate the representation performance of a neural network:
\begin{equation}
    E = \frac{1}{2} \sum_{x} [ I(x) - \sum_i w_i \phi_i (x) ]^2 + \lambda \sum_i S(\frac{w_i}{\sigma})
\end{equation}
Here, we assume that neural networks learn to find a set of basis functions $\{\phi_i\}$ that represents each image in terms of a linear combination, and the coefficients $\{w_i\}$ together reflect the neural network structure features.
The $S(x)$ in the second term is a nonlinear function assessing the sparseness of the network, with a scaling constant $\sigma$.

Learning is accomplished by minimizing the total cost $E$ 
utilizing $ \frac{\partial{E}}{\partial{w_i}} $ and 
$ \frac{\partial{E}}{\partial{\phi_i}} $, i.e.:
\begin{equation}
    \frac{\partial{E}}{\partial{w_i}} 
    = - \sum_{x} \phi_i ( I - \sum_j w_j \phi_j )+ \frac{\lambda}{\sigma} S^{\prime}(\frac{w_i}{\sigma})
\end{equation}
\begin{equation}
    \frac{\partial{E}}{\partial{\phi_i}} 
    = - \sum_x w_i (I - \sum_j w_j \phi_j )
\end{equation}
Intuitively, the algorithm is seeking a set of $\{w_i\}$ and $\{\phi_i\}$ to tolerate sparsification as well as reaching minimum reconstruction error, which also reflects that a given neural network can adapt to the changes brought by parameter pruning.

Next, let us consider the pruning-during-training process, suppose the parameters are pruned through mask, therefore they can be re-activated during the process, see section 3.2 for details.
We assume a set of $\{a_k\}$ as the unmasked parameters is chosen at next pruning stage, i.e. stage $t+1$, and $\sum_i w_i \phi_i (x)$ represents the neural network with unmasked parameters $\{w_i\}$ at current stage $t$, 
s.t. $\{a_k\} = (\{w_i\} \setminus \{b_h\}) \cup \{c_j\}$,
where $\{b_h\}$ are pruned and $\{c_j\}$ 
are re-activated at stage $t+1$.
Then we should have:
\begin{align}
    E_{t+1} 
    &= \frac{1}{2} \sum_{x} [ I(x) - \sum_k a_k \phi_k (x) ]^2 + \lambda \sum_k S(\frac{a_k}{\sigma}) \notag \\
    &= \frac{1}{2} \sum_{x} [ I(x) - \sum_i w_i \phi_i (x) + \sum_h b_h \phi_h (x) - \sum_j c_j \phi_j (x) ]^2 + \lambda \sum_k S(\frac{a_k}{\sigma}) \notag \\
    &= \frac{1}{2} \sum_{x} ( I(x) - \sum_i w_i \phi_i (x) )^2 + 
    \frac{1}{2} \sum_{x} (\sum_h b_h \phi_h (x) - \sum_j c_j \phi_j (x))^2 \notag \\
    &+ \sum_{x} (I(x) - \sum_i w_i \phi_i (x))(\sum_h b_h \phi_h (x) - \sum_j c_j \phi_j (x)) + \lambda \sum_k S(\frac{a_k}{\sigma}) 
\end{align}
Intuitively, the first term in equation (5) aims to keep updating the network to fit the given input data, and the second term means that the re-activated part should compensate for the lost information, while the third term prefers a zero covariance between two kinds of biases.
We further derive the updating rules as below:
\begin{equation}
\frac{\partial{E_{t+1}}}{\partial{w_k}} 
= - \sum_{x} \phi_k ( I - \sum_i w_i \phi_i ) 
- \sum_{x} \phi_k ( \sum_h b_h \phi_h - \sum_j c_j \phi_j ) 
+ \frac{\lambda}{\sigma} S^{\prime}(\frac{w_k}{\sigma})
\end{equation}
\begin{equation}
\frac{\partial{E_{t+1}}}{\partial{c_k}} 
= - \sum_{x} \phi_k ( \sum_h b_h \phi_h - \sum_j c_j \phi_j ) 
- \sum_{x} \phi_k ( I - \sum_i w_i \phi_i ) 
+ \frac{\lambda}{\sigma} S^{\prime}(\frac{c_k}{\sigma})
\end{equation}
\begin{equation}
    \frac{\partial{E_{t+1}}}{\partial{\phi_k}} 
    = - \sum_x a_k (I - \sum_i w_i \phi_i ) - \sum_x a_k ( \sum_h b_h \phi_h - \sum_j c_j \phi_j )
\end{equation}
Compared to equation (3) and (4), we argue that the differences reveal the underlying cause of \textbf{Neural Burden} phenomenon in a sense.
We further examine this phenomenon through experiment in the next section.






\begin{figure}[t]
    \includegraphics[width=\textwidth]{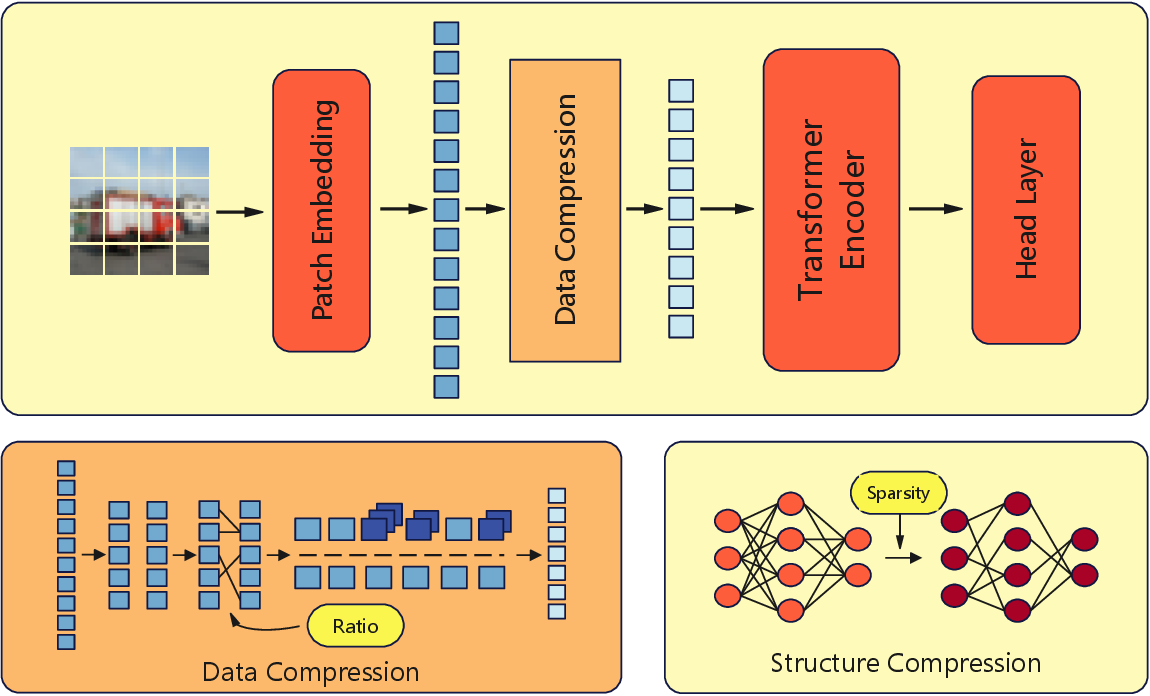}
    \caption{Overview of our proposed joint compression framework. 
    We use a bipartite matching method to perform token merging on the data before sent to the transformer encoder, thereby compressing the data. 
    Simultaneously, a dynamic pruning method is employed throughout the training process. 
    To validate the effectiveness of the framework, both algorithms are implemented using relatively simple approaches commonly used in their respective fields.} 
    \label{fig1}
\end{figure}



\subsection{A Simple Enhancement Solution for Pruning}

Research in neuroscience has indicated that the brain performs better under low-load conditions compared to high-load situations\cite{effectsofvisualworkingmemoryload}.
Inspired by this, in order to alleviate the Neural Burden and the performance degradation of pruning methods, we propose adding a simple plug-and-play data compression module before pruning to pre-compress information sent to neural networks, thereby reducing the neuron load.
Note that this can be used for a wide range of pruning methods.
The overall framework is then shown in Figure \ref{fig1}. 
We use Vision Transformer (ViT) as the base model, where images are patchified into token sequences.
These tokens can be viewed as relatively independent information units, and naturally, some of these tokens might be highly similar or redundant for the final prediction. 
Thus, compressing the information sent into the neural network at this level is reasonable, thereby effectively reducing the load of the neurons.
A gradual iterative pruning method is adopted here throughout the training process to achieve the pre-setted sparsity. 

\subsubsection{Single-Pass Token Merging for Data Compression}

We aim to preprocess input data through compression. 
Considering that compressed tokens inevitably lose information, we opted for a token merging approach to combine similar tokens and preserve as much information as possible. 
The original token merging algorithm performs this operation between the multi-head self-attention (MHA) and the feed-forward neural network (FFN) in each transformer encoder, reusing the $K\in R^{n\times d}$ matrix from the MHA calculations to measure the similarity between tokens.
In our experiments, we observed that in a globally sparsified neural network, the sparsity levels of different subnetworks are roughly the same. 
Thus, as shown in Figure 1, we opted to apply the token merging operation between the patch embedding and the transformer encoder in Vision Transformers (ViTs).
To measure token similarity, we referenced the original method of the token merging algorithm and employed the bipartite soft matching algorithm for matching, obtaining similarity scores by directly computing the dot product matrix between token vectors.
Since this operation is performed only once, the additional computational overhead is acceptable. 
Our algorithm is as follows:
\begin{description}
    \item[step1] Divide the token sequence into two equal-sized sets, $A$ and $B$.
    \item[step2] Compute similarity between two sets and retain the most similar pairs.
    \item[step3] Merge the most similar tokens.
    \item[step4] Combine the two sets into a single token sequence.
\end{description}

\subsubsection{Persistent Dynamic Weight Pruning Strategy}

The training of neural network models can be likened to the establishment of neuronal connections in the brain.
In our pruning-during-training setting, we continuously prune weights throughout all training epochs. 
We adopt a relatively simple method for dynamic weight pruning. 
For a neural network to be pruned, we impose a mask on all weights to sparsify the weight matrix. 
\begin{equation}
    \tilde{w}=w\odot m    
\end{equation}
During training, to ensure the model reaches the given sparsity $r$, we calculate the targeted sparsity $r_t$ after a fixed number of iterations $t$. 
\begin{equation}
    r_t=r-r(1-\frac{t}{iterations_{all}})^3
\end{equation}
\begin{equation}
    \dot{k_t}=len(w)*(1-r_t)
\end{equation}
\begin{equation}
    threshold_t=\text{$\dot{k_t}$-th largeset $ImportanceScore(w)$}
\end{equation}
Then, by updating the mask with the threshold, the model reaches a new intermediate sparsity,
\begin{equation}
    m_{t}=gate(ImportanceScore(w), threshold_t)
\end{equation}
where $m_0$ is a mask with all values of 1.
Note that, applying such updating strategy allows to recover from “errors”, i.e. parameters during pruning can become activated again.



\section{Experiments}

In this section, we conduct experiments to examine the proposed concept of Neural Burden and the effectiveness of our enhancement method for pruning. 
To more intuitively present our results, we only consider highly sparse models here to avoid the effects of over-parameterization.
All experiments were conducted using the Python framework on a single NVIDIA 3090 GPU, with a fixed random seed to ensure the reproducibility of the results.

\subsection{Setup}


\subsubsection{Datasets}
We adopt MNIST and CIFAR-10 for our experiments. 
The MNIST training set contains 60,000 images, while the test set contains 10,000 images, with the size of 32x32x1 and labels divided into 10 categories. 
CIFAR-10 contains 60,000 images sized 32x32x3 with a training-test split ratio of 5:1, which are also divided into 10 categories. 
We preprocessed the images in CIFAR-10 dataset to a size of 224x224x3 to enlarge the number of input tokens, and set the ViT patch size to 16.

\subsubsection{Architecture}
For the analysis experiments conducted on MNIST, we selected an MLP for simplicity to facilitate the display of relevant neural burden indicators. 
For the validation experiments on CIFAR-10, we used a four-layer ViT as the base model for the compression algorithm to verify the effectiveness and applicability of our method.

\subsubsection{Experimental Details}
In the analysis experiments using MLP, we inspected networks with the same sparsity level. 
The indicators included weights and gradients recorded at equal intervals (reflecting the instantaneous state and dynamic characteristics of the model), or the cumulative mean at fixed training intervals (reflecting the overall change trend of the model), to analyze the impact of neural burden on sparse models. 
In the validation experiments using ViT, we examined the improvement effect of data compression under different settings.

\subsection{Neural burden affects the training of highly sparse neural networks}
We aim to examine whether neural networks exhibit neural burden phenomenon similar to the brain's information propagation mechanism. 
Therefore, we trained an MLP on MNIST with network sparsity levels (i.e., the proportion of zero values in parameters) set at 0.7, 0.8, and 0.9. 
We conducted three comparative strategies:
(i) Iterative Pruning, which dynamically prunes the network throughout the training process, eventually achieving the given sparsity level;
(ii) Optimal Subnetwork, we use the parameter mask obtained from the iterative pruning at initialization to directly obtain the subnetwork, which will not be pruned in the subsequent training;
(iii) Random Subnetwork, which trains the network after randomly pruning an initialized dense network to the given sparsity level.

\begin{figure}[!t]
\centering
	\subfloat[gradient]{\includegraphics[width=0.8\linewidth]{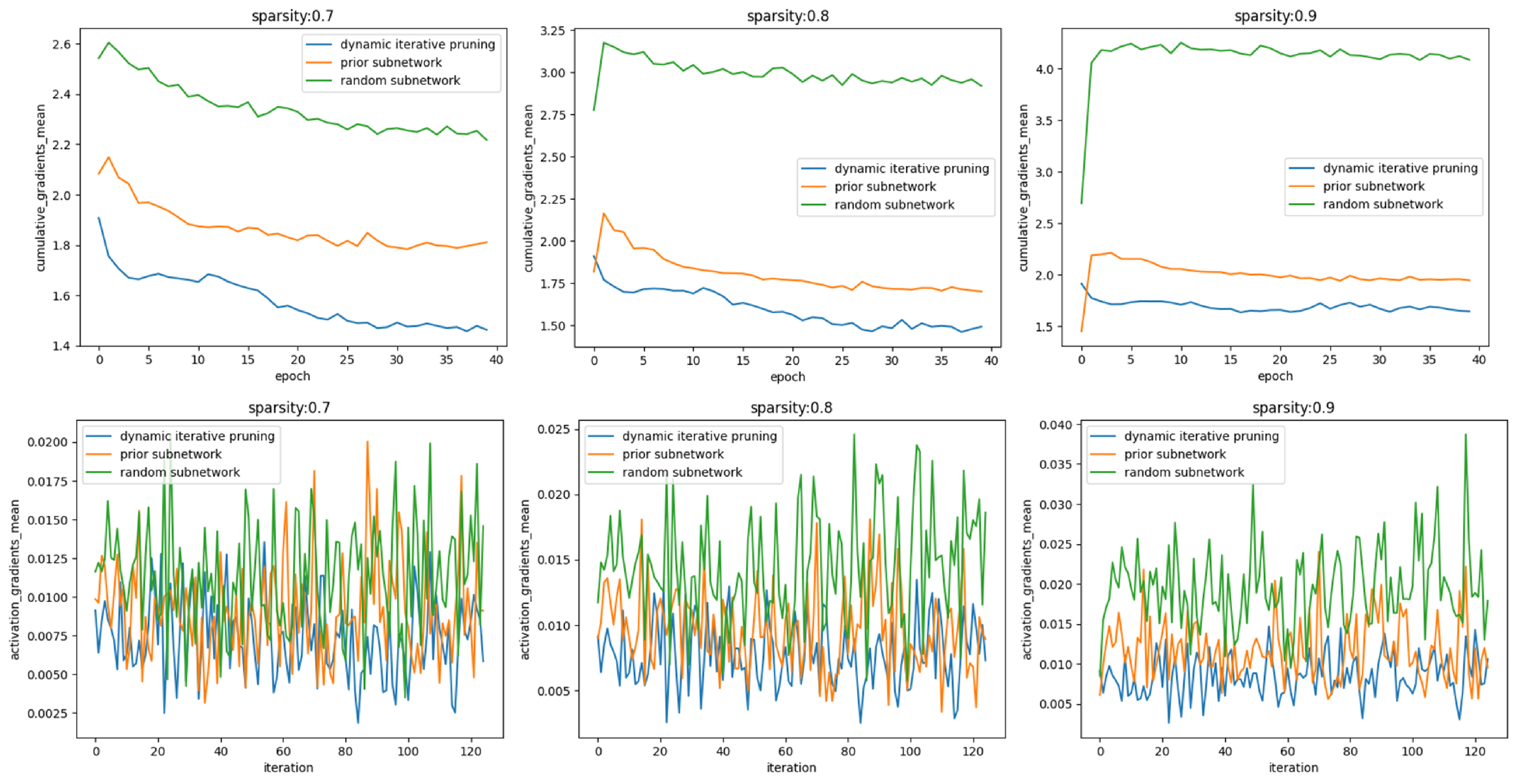}} \\
	\subfloat[weight]{\includegraphics[width=0.8\linewidth]{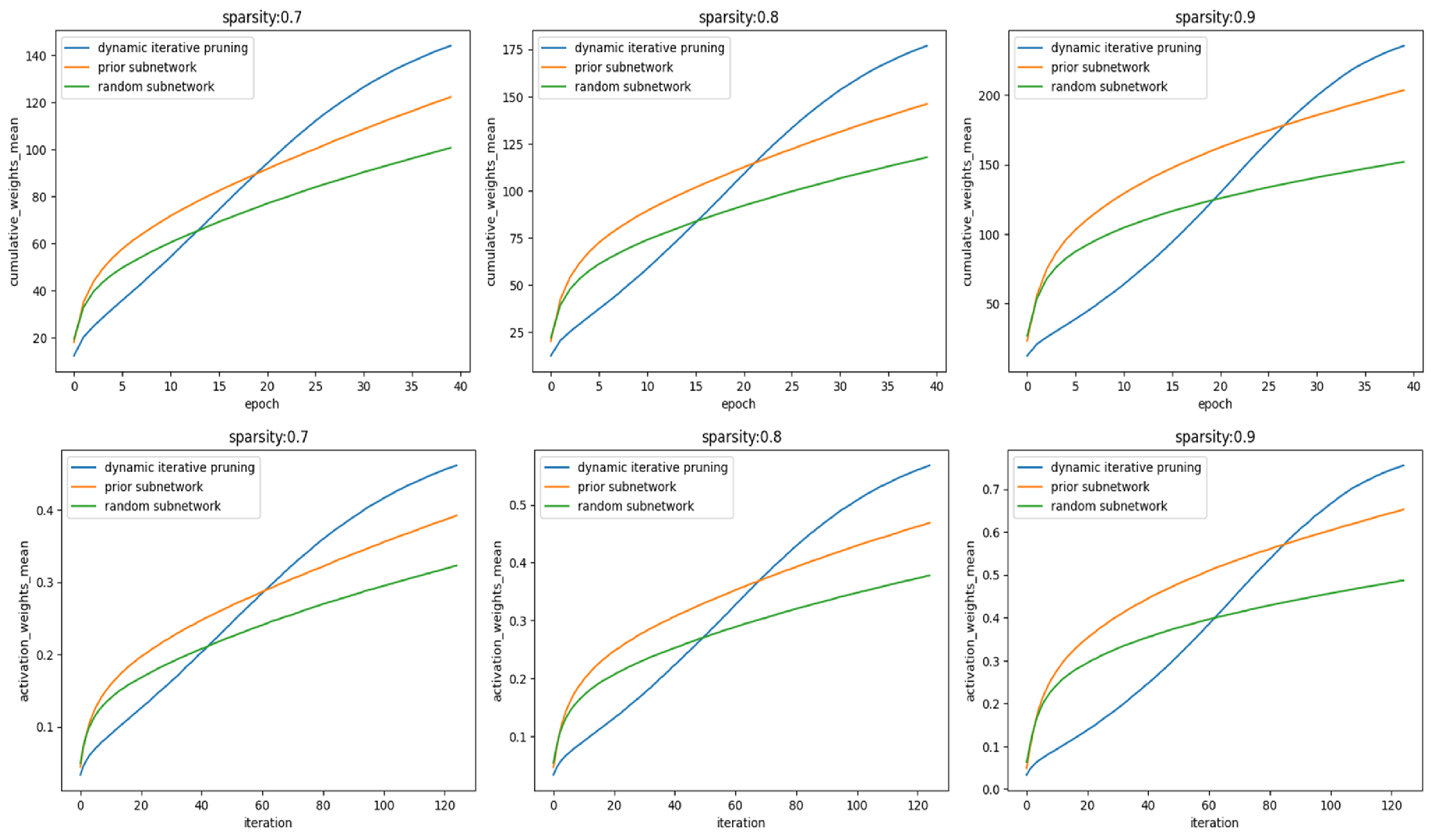}}
\caption{For each subgraph, the first row shows the cumulative mean of gradients/weights for neurons recorded in each epoch, and the second row shows the mean of gradients/weights for neurons recorded on every 100 steps. The blue curve represents the dynamic iterative pruning experiment, the yellow curve represents the experiment of training from scratch using the optimal subnetwork, and the green curve represents the experiment of training from scratch using a random subnetwork.}
\label{neural burden}
\end{figure}

As shown in Fig.2(a), the gradient magnitudes of the three networks follow the same order across different sparsity levels, namely iterative pruning \textless optimal subnetwork \textless random subnetwork.
Notably, the gradient magnitudes of the iterative pruning network are the lowest at most of the time.
We argue that this is because during the pruning process, the artificial network neurons need to recover the lost information previously preserved by the pruned neurons, while simultaneously fitting the training data, resulting in an excessive load. 
Thus, following a similar manner of high-load neurons observed in neuroscience research (for example, \cite{macnamara2012electrocortical} reported decreased face-related N170 amplitudes under high load), the artificial neurons decrease their attention upon some information which may include task-relevant information, leading to a reduction in gradients.
These results indicate that the Neuron Burden phenomenon that we proposed indeed exists.

Moreover, we inspected the norm of weights during training.
In the case of gradually reducing neurons in the network, there are more neurons carrying information in the early stages, so the activation of a single neuron is relatively small; in the later stages of training, the remaining neurons have to gradually bear the effective activation previously handled by the pruned neurons, thus the activation is excessive and the importance of the neurons is relatively prominent. 
This also coincides with the compensation observed in \cite{barrett2016optimal}.

\subsection{The effect of data compression on pruning}

The phenomenon of neural burden suggests that highly sparse networks have a limited capacity to handle information. 
Existing research on the connection structure and information transmission mechanisms in brain neurons reveals that information may need to be compressed when propagated through the neural network during learning. 
We decided to compress input data to investigate whether this method can effectively reduce performance loss caused by neural burden during pruning a model.

We observed the performance of iteratively pruned ViTs on CIFAR-10 with sparsity levels of {0.75, 0.8, 0.85, 0.9, 0.95} under different data compression ratios {0.2, 0.4, 0.6, 0.8}, with embedding size ranging from {384, 768}. 
The results of the experiments are organized in Table 1 and Table 2.

According to the results in Table 1 and Table 2, our proposed solution improved the performance of highly sparse pruned networks in most experimental configurations. 
This indicates that for neural networks to be highly pruned, pre-compressing the data can effectively reduce the load during the learning process, thereby enhancing performance.

\begin{table}[!ht]
    \caption{Test accuracy under model embedding size of 384.}
    \centering
    \resizebox{0.95\textwidth}{!}{
    \begin{tabular}{c|ccccc|ccccc|ccccc|ccccc}
    \hline 
        \multicolumn{21}{c}{Experiment Setup: Embed-Dim=384, Head-Num=3, Epoch=40} \\ \hline
        Learning Rate & \multicolumn{5}{c}{1.25e-3} & \multicolumn{5}{c}{2.5e-3} & \multicolumn{5}{c}{5e-3} & \multicolumn{5}{c}{1e-2} \\ \hline
        \diagbox{Merge}{Sparsity} & 0.75 & 0.80 & 0.85 & 0.90 & 0.95 & 0.75 & 0.80 & 0.85 & 0.90 & 0.95 & 0.75 & 0.80 & 0.85 & 0.90 & 0.95 & 0.75 & 0.80 & 0.85 & 0.90 & 0.95 \\ \hline

        0 & 75 & 74.33 & 74.46 & 74.03 & 72.82 & 76.93 & 76.26 & 76.09 & 76.04 & 75.26 & 77.41 & 77.27 & 76.83 & 76.84 & 76.44 & 78.02 & 77.75 & 76.76 & 76.76 & 76.6 \\ \hline
        0.2 & 74.59 & \textbf{74.4} & 73.96 & 73.43 & 72.4 & \textbf{77.09} & \textbf{76.59} & 76.06 & 75.75 & \textbf{75.41} & \textbf{77.72} & \textbf{77.59} & \textbf{77.16} & \textbf{77.12} & \textbf{76.63} & \textbf{78.35} & \textbf{77.93} & \textbf{78} & \textbf{77.74} & \textbf{77.53} \\ 
        0.4 & 74.64 & \textbf{74.5} & 74 & \textbf{74.04} & \textbf{73.27} & 76.29 & \textbf{76.34} & 75.49 & 75.55 & 74.64 & \textbf{78.19} & \textbf{77.32} & \textbf{77.44} & \textbf{76.99} & \textbf{76.48} & \textbf{78.23} & 77.57 & \textbf{78.28} & \textbf{77.65} & \textbf{77.13} \\ 
        0.6 & 74.55 & 74.06 & \textbf{74.61} & 73.18 & \textbf{72.88} & 76.28 & 75.69 & 75.98 & 75.6 & 74.92 & \textbf{77.79} & \textbf{77.35} & \textbf{77.55} & \textbf{76.87} & \textbf{76.5} & \textbf{78.2} & \textbf{78.28} & \textbf{77.98} & \textbf{77.59} & \textbf{77.01} \\ 
        0.8 & 74.7 & \textbf{74.69} & 73.96 & \textbf{74.08} & \textbf{72.95} & 76.15 & 75.55 & 75.21 & 75.68 & 73.98 & \textbf{77.56} & 77.26 & \textbf{76.92} & 76.81 & \textbf{76.77} & 77.37 & 77.01 & \textbf{76.93} & 76.46 & 75.9 \\ 

        \hline \hline
        \multicolumn{21}{c}{Experiment Setup: Embed-Dim=384, Head-Num=3, Epoch=60} \\ \hline 
        Learning Rate & \multicolumn{5}{c}{1.25e-3} & \multicolumn{5}{c}{2.5e-3} & \multicolumn{5}{c}{5e-3} & \multicolumn{5}{c}{1e-2} \\ \hline
        \diagbox{Merge}{Sparsity} & 0.75 & 0.80 & 0.85 & 0.90 & 0.95 & 0.75 & 0.80 & 0.85 & 0.90 & 0.95 & 0.75 & 0.80 & 0.85 & 0.90 & 0.95 & 0.75 & 0.80 & 0.85 & 0.90 & 0.95 \\ \hline         

        0 & 76.19 & 75.94 & 75.8 & 75.15 & 74.36 & 77.16 & 77.46 & 77.27 & 76.87 & 76.9 & 79.26 & 79.36 & 79.02 & 78.87 & 78.37 & 80.87 & 80.21 & 79.87 & 80.26 & 79.48 \\ \hline   
        0.2 & 76.1 & 75.74 & 75.8 & 74.94 & \textbf{74.62} & \textbf{78.32} & \textbf{77.74} & \textbf{77.54} & \textbf{77.38} & \textbf{77.41} & 79.11 & 78.82 & \textbf{79.34} & 78.65 & 78.28 & 80.39 & \textbf{80.69} & \textbf{80.29} & \textbf{80.53} & \textbf{80.03} \\ 
        0.4 & \textbf{76.39} & \textbf{76.05} & 75.35 & \textbf{75.37} & \textbf{74.57} & \textbf{77.74} & \textbf{77.87} & 76.92 & \textbf{77.27} & 76.48 & \textbf{79.46} & 78.83 & 79.02 & 78.58 & \textbf{78.51} & 80.75 & 79.6 & 79.82 & 79.87 & \textbf{80.01} \\ 
        0.6 & 76.16 & \textbf{75.96} & 75.02 & \textbf{75.29} & 74.08 & \textbf{77.78} & 77.28 & 76.64 & 76.68 & 76.26 & \textbf{80.02} & 79.32 & \textbf{79.17} & \textbf{79.47} & \textbf{79.14} & 80.16 & 80.03 & \textbf{79.9} & \textbf{80.3} & \textbf{79.65} \\ 
        0.8 & \textbf{76.52} & \textbf{76.1} & 75.33 & \textbf{75.18} & \textbf{74.77} & \textbf{77.87} & \textbf{77.81} & 77.22 & \textbf{77.17} & \textbf{77.25} & 78.94 & 78.44 & \textbf{79.23} & 78.4 & \textbf{78.67} & 79.66 & 79.5 & 79.81 & 79.04 & 78.85 \\ 

        \hline \hline
        \multicolumn{21}{c}{Experiment Setup: Embed-Dim=384, Head-Num=6, Epoch=40} \\ \hline 
        Learning Rate & \multicolumn{5}{c}{1.25e-3} & \multicolumn{5}{c}{2.5e-3} & \multicolumn{5}{c}{5e-3} & \multicolumn{5}{c}{1e-2} \\ \hline
        \diagbox{Merge}{Sparsity} & 0.75 & 0.80 & 0.85 & 0.90 & 0.95 & 0.75 & 0.80 & 0.85 & 0.90 & 0.95 & 0.75 & 0.80 & 0.85 & 0.90 & 0.95 & 0.75 & 0.80 & 0.85 & 0.90 & 0.95 \\ \hline
        
        0 & 76.48 & 75.9 & 75.63 & 75.47 & 74.25 & 76.98 & 76.69 & 76.42 & 76.44 & 75.51 & 77.52 & 77.34 & 76.64 & 76.88 & 76.33 & 77.74 & 78.26 & 77.47 & 77.48 & 76.94 \\ \hline
        0.2 & 75.79 & 75.7 & \textbf{75.69} & \textbf{75.52} & \textbf{74.41} & \textbf{77.01} & 76.3 & 75.87 & 75.92 & 74.9 & 77.31 & 77.1 & \textbf{76.83} & 76.78 & 76.08 & \textbf{78.33} & 77.71 & \textbf{77.88} & 77.13 & \textbf{77.47} \\ 
        0.4 & 75.72 & 75.29 & \textbf{75.75} & 75.22 & 73.73 & 76.9 & 76.69 & \textbf{76.43} & 76.29 & \textbf{75.89} & \textbf{78.08} & \textbf{77.68} & \textbf{77.2} & \textbf{77.33} & \textbf{76.52} & \textbf{78.62} & 78.19 & \textbf{77.94} & \textbf{77.57} & \textbf{77.08} \\ 
        0.6 & 75.86 & 75.66 & 75.44 & 75.35 & \textbf{74.48} & 76.82 & 76.48 & \textbf{76.52} & 76.27 & \textbf{75.84} & \textbf{78} & \textbf{78.09} & \textbf{77.35} & \textbf{77.77} & \textbf{77.01} & \textbf{78.43} & 77.99 & \textbf{77.57} & 77.4 & 76.92 \\ 
        0.8 & 75.51 & 75.09 & 75.46 & 75 & 73.45 & 76.95 & 76.29 & 76.14 & 76.12 & 75.38 & \textbf{77.75} & 77.25 & \textbf{76.87} & 76.2 & \textbf{76.7} & \textbf{78.07} & 77.88 & \textbf{77.88} & 76.93 & 76.7 \\ 

        \hline \hline
        \multicolumn{21}{c}{Experiment Setup: Embed-Dim=384, Head-Num=6, Epoch=60} \\ \hline 
        Learning Rate & \multicolumn{5}{c}{1.25e-3} & \multicolumn{5}{c}{2.5e-3} & \multicolumn{5}{c}{5e-3} & \multicolumn{5}{c}{1e-2} \\ \hline
        \diagbox{Merge}{Sparsity} & 0.75 & 0.80 & 0.85 & 0.90 & 0.95 & 0.75 & 0.80 & 0.85 & 0.90 & 0.95 & 0.75 & 0.80 & 0.85 & 0.90 & 0.95 & 0.75 & 0.80 & 0.85 & 0.90 & 0.95 \\ \hline
        
        0 & 76.59 & 76.19 & 75.83 & 75.31 & 75.53 & 78.02 & 78.56 & 77.59 & 77.11 & 76.41 & 79.29 & 79.21 & 79.59 & 79.24 & 77.83 & 80.27 & 80.57 & 79.25 & 79.82 & 79.69 \\ \hline
        0.2 & 76.39 & 75.53 & 75.68 & 75.3 & 74.26 & \textbf{78.42} & 77.48 & 77.43 & 76.99 & \textbf{76.7} & 78.38 & 78.33 & 78.36 & 78.82 & 77.81 & 80.11 & 80.4 & \textbf{80.43} & 79.73 & 79.36 \\ 
        0.4 & 76.55 & \textbf{76.2} & 75.56 & 75.18 & 74.75 & \textbf{78.1} & 77.67 & 77.55 & \textbf{77.43} & \textbf{76.66} & 79.29 & 79.05 & 79.09 & 78.85 & \textbf{78.09} & \textbf{80.52} & 79.67 & \textbf{79.92} & 79.74 & 79.67 \\ 
        0.6 & 76.58 & 76.16 & \textbf{75.98} & \textbf{75.75} & 74.7 & 77.92 & 78.08 & \textbf{77.93} & 77.07 & \textbf{76.65} & \textbf{79.69} & \textbf{79.81} & \textbf{79.61} & 78.8 & \textbf{78.27} & 80.03 & 80.43 & \textbf{79.64} & 79.77 & 79.47 \\ 
        0.8 & 76.29 & \textbf{76.36} & 75.49 & 75.21 & 74.78 & \textbf{78.37} & 77.97 & \textbf{78.15} & \textbf{77.3} & \textbf{76.98} & 78.79 & 79.11 & 78.8 & 78.72 & 77.71 & \textbf{80.48} & 79.9 & \textbf{80.41} & 79.74 & 79.18 \\ 

        \hline
    \end{tabular}}
\end{table}

\begin{table}[!ht]
    \caption{Test accuracy under model embedding size of 768.}
    \centering
    \resizebox{0.95\textwidth}{!}{
    \begin{tabular}{c|ccccc|ccccc|ccccc|ccccc}
    \hline 
        \multicolumn{21}{c}{Experiment Setup: Embed-Dim=768, Head-Num=3, Epoch=40} \\ \hline
        Learning Rate & \multicolumn{5}{c}{1.25e-3} & \multicolumn{5}{c}{2.5e-3} & \multicolumn{5}{c}{5e-3} & \multicolumn{5}{c}{1e-2} \\ \hline
        \diagbox{Merge}{Sparsity} & 0.75 & 0.80 & 0.85 & 0.90 & 0.95 & 0.75 & 0.80 & 0.85 & 0.90 & 0.95 & 0.75 & 0.80 & 0.85 & 0.90 & 0.95 & 0.75 & 0.80 & 0.85 & 0.90 & 0.95 \\ \hline

        0 & 79.66 & 79.18 & 79.51 & 78.59 & 77.9 & 79.42 & 79.02 & 78.93 & 78.71 & 78.76 & 79.73 & 78.95 & 78.99 & 79.48 & 78.9 & 61.32 & 59.81 & 72.59 & 78.41 & 76.72 \\ \hline
        0.2 & 78.87 & 78.67 & 78.89 & 78.34 & \textbf{78.21} & 78.51 & 78.75 & 78.53 & 78.53 & 78.21 & 79.58 & \textbf{79.52} & \textbf{79.15} & 79.01 & \textbf{79.33} & \textbf{75.54} & \textbf{77.11} & \textbf{78.18} & 74.42 & \textbf{78.28} \\ 
        0.4 & 79.38 & 78.9 & 78.65 & \textbf{78.76} & \textbf{78.07} & 79.31 & \textbf{79.49} & \textbf{79} & \textbf{79.3} & \textbf{79.02} & \textbf{79.96} & \textbf{79.62} & \textbf{79.82} & 78.77 & 78.68 & 54.19 & \textbf{78.06} & 68.29 & 75.43 & \textbf{77.96} \\ 
        0.6 & 79 & 78.53 & 79.04 & \textbf{78.89} & \textbf{78.2} & 79.05 & \textbf{79.36} & \textbf{78.98} & 78.3 & 78.55 & \textbf{79.97} & \textbf{79.53} & \textbf{79.7} & 78.7 & \textbf{79.3} & 52.27 & \textbf{72.3} & 57.03 & 73.16 & \textbf{78.61} \\ 
        0.8 & 79.2 & 78.85 & 78.94 & 78.18 & 77.9 & \textbf{79.67} & \textbf{79.24} & \textbf{79.09} & \textbf{78.93} & 78.4 & 79.29 & \textbf{79} & \textbf{79.24} & 79.01 & 78.59 & \textbf{74.14} & \textbf{70.98} & \textbf{77.11} & 74.47 & \textbf{77.49} \\ 

        \hline \hline
        \multicolumn{21}{c}{Experiment Setup: Embed-Dim=768, Head-Num=3, Epoch=60} \\ \hline 
        Learning Rate & \multicolumn{5}{c}{1.25e-3} & \multicolumn{5}{c}{2.5e-3} & \multicolumn{5}{c}{5e-3} & \multicolumn{5}{c}{1e-2} \\ \hline
        \diagbox{Merge}{Sparsity} & 0.75 & 0.80 & 0.85 & 0.90 & 0.95 & 0.75 & 0.80 & 0.85 & 0.90 & 0.95 & 0.75 & 0.80 & 0.85 & 0.90 & 0.95 & 0.75 & 0.80 & 0.85 & 0.90 & 0.95 \\ \hline
        
        0 & 79.99 & 80.02 & 79.25 & 79.2 & 78.85 & 80.04 & 79.72 & 79.58 & 79.47 & 79.06 & 80.74 & 80.44 & 80.37 & 80.25 & 79.61 & 67.93 & 67.28 & 70.03 & 71.66 & 70.15 \\ \hline
        0.2 & 79.97 & \textbf{80.25} & 78.76 & 78.99 & 78.26 & 79.18 & 79.64 & \textbf{79.61} & 79.07 & 79.01 & \textbf{80.87} & 80.42 & 80.01 & 79.93 & 79.58 & \textbf{68.3} & \textbf{68.33} & 64.58 & 71.1 & \textbf{72.32} \\ 
        0.4 & 79.51 & 79.84 & \textbf{79.28} & \textbf{79.34} & 77.9 & \textbf{80.43} & 79.69 & \textbf{79.88} & 78.27 & \textbf{79.19} & 80.6 & 79.76 & 80.33 & \textbf{80.4} & \textbf{79.92} & 62.75 & 67.04 & \textbf{71.35} & 71.1 & 69.93 \\ 
        0.6 & 79.51 & 79.67 & \textbf{79.6} & 79.01 & 78.49 & 79.39 & \textbf{79.98} & \textbf{80.09} & \textbf{79.75} & \textbf{79.23} & 80.18 & \textbf{81.26} & \textbf{80.42} & 80.13 & \textbf{80.2} & 63.62 & 63.33 & 67.61 & 70.39 & \textbf{73} \\ 
        0.8 & 79.62 & 79.6 & 78.84 & \textbf{79.54} & 78.52 & 80.04 & 79.67 & \textbf{79.76} & 79.21 & \textbf{79.35} & 80.26 & 79.91 & 80.2 & 79.84 & \textbf{79.96} & 63.15 & \textbf{71.25} & 67.5 & 69.11 & \textbf{70.67} \\ 
 
        \hline \hline
        \multicolumn{21}{c}{Experiment Setup: Embed-Dim=768, Head-Num=6, Epoch=40} \\ \hline 
        Learning Rate & \multicolumn{5}{c}{1.25e-3} & \multicolumn{5}{c}{2.5e-3} & \multicolumn{5}{c}{5e-3} & \multicolumn{5}{c}{1e-2} \\ \hline
        \diagbox{Merge}{Sparsity} & 0.75 & 0.80 & 0.85 & 0.90 & 0.95 & 0.75 & 0.80 & 0.85 & 0.90 & 0.95 & 0.75 & 0.80 & 0.85 & 0.90 & 0.95 & 0.75 & 0.80 & 0.85 & 0.90 & 0.95 \\ \hline       

        0 & 79.01 & 79.08 & 78.85 & 78.75 & 77.9 & 79.63 & 79.85 & 78.83 & 78.89 & 79.04 & 79.31 & 80.21 & 79.6 & 79.09 & 79.01 & 63.39 & 65.6 & 76.14 & 78.12 & 77.25 \\ \hline
        0.2 & \textbf{79.62} & \textbf{79.39} & \textbf{79.33} & \textbf{78.88} & 77.72 & 79.52 & \textbf{79.92} & \textbf{79.73} & \textbf{79.51} & \textbf{79.12} & \textbf{79.73} & 80.09 & \textbf{79.65} & \textbf{79.58} & \textbf{79.06} & 57.28 & \textbf{66.39} & 64.24 & 75.9 & 76.78 \\ 
        0.4 & \textbf{79.31} & \textbf{79.53} & \textbf{79.43} & \textbf{79.21} & \textbf{78.71} & \textbf{80.39} & \textbf{79.98} & \textbf{79.76} & \textbf{79.78} & \textbf{79.37} & \textbf{79.48} & 79.31 & 79.39 & \textbf{79.11} & \textbf{79.07} & 56.34 & \textbf{65.72} & \textbf{76.34} & \textbf{78.68} & \textbf{79.14} \\ 
        0.6 & \textbf{79.73} & \textbf{79.66} & \textbf{79.63} & \textbf{79.51} & \textbf{78.76} & \textbf{80.02} & 79.82 & \textbf{79.63} & \textbf{79.11} & \textbf{79.39} & \textbf{79.91} & 79.88 & \textbf{79.82} & \textbf{79.37} & \textbf{79.24} & 61.55 & 62.01 & 60.22 & \textbf{78.66} & \textbf{77.43} \\ 
        0.8 & \textbf{79.45} & \textbf{79.38} & \textbf{79.03} & \textbf{78.99} & \textbf{78.26} & \textbf{79.69} & 79.46 & \textbf{79.46} & \textbf{79.24} & 78.64 & 79.24 & 79.4 & 79.43 & 78.8 & 78.53 & 57.91 & 62.18 & 63.06 & 65.95 & 68.61 \\ 

        \hline \hline
        \multicolumn{21}{c}{Experiment Setup: Embed-Dim=768, Head-Num=6, Epoch=60} \\ \hline 
        Learning Rate & \multicolumn{5}{c}{1.25e-3} & \multicolumn{5}{c}{2.5e-3} & \multicolumn{5}{c}{5e-3} & \multicolumn{5}{c}{1e-2} \\ \hline
        \diagbox{Merge}{Sparsity} & 0.75 & 0.80 & 0.85 & 0.90 & 0.95 & 0.75 & 0.80 & 0.85 & 0.90 & 0.95 & 0.75 & 0.80 & 0.85 & 0.90 & 0.95 & 0.75 & 0.80 & 0.85 & 0.90 & 0.95 \\ \hline
        
        0 & 79.91 & 79.8 & 79.17 & 79.14 & 78.2 & 80.49 & 80.3 & 80.07 & 80.26 & 79.26 & 80.72 & 80.41 & 81.37 & 79.44 & 79.99 & 70.87 & 74.05 & 74.43 & 72.63 & 73.98 \\ \hline
        0.2 & \textbf{79.96} & \textbf{79.96} & \textbf{79.73} & \textbf{79.29} & \textbf{78.7} & 80.41 & 79.84 & 79.7 & 80.03 & 79.13 & 80.71 & 80.22 & 79.83 & \textbf{80.55} & 79.97 & 70.1 & 70.7 & 73.09 & \textbf{74.1} & \textbf{75.8} \\ 
        0.4 & \textbf{80} & 79.55 & \textbf{79.63} & \textbf{79.32} & \textbf{78.59} & 80.34 & 79.71 & 79.8 & 79.9 & 79.26 & 80.07 & 79.89 & 79.86 & \textbf{79.66} & 79.38 & 68.67 & 69.5 & 69.07 & \textbf{73.06} & \textbf{74.25} \\ 
        0.6 & \textbf{80.36} & \textbf{80.07} & \textbf{80.14} & \textbf{79.23} & \textbf{79.09} & 80.46 & 80.1 & \textbf{80.26} & 79.99 & \textbf{79.54} & 79.5 & 79.87 & 79.2 & \textbf{79.71} & 79.67 & 66.13 & 68.64 & 72.99 & 71.38 & 73.59 \\ 
        0.8 & \textbf{80.04} & 79.62 & \textbf{79.94} & \textbf{79.62} & \textbf{78.39} & 79.81 & 80.01 & 79.77 & 79.63 & 78.89 & 79.27 & 79.85 & 79.57 & \textbf{79.96} & 79.48 & 68.37 & 71.69 & 72.17 & \textbf{75.25} & 72.48 \\ 
        
        \hline
    \end{tabular}}
\end{table}

\section{Discussion}
In addition to the above content, there are several directions to explore in the future. 
The data compression method used in current experiments is limited to the token level and the similarity calculation. 
Although this method simplifies input information by merging tokens, it still results in the loss of valuable information. 
Future research could explore data encoding at the embedding level to build a more effective data utilization mechanism, or could explore token merging method via mutual information, etc. 
Moreover, neuroscience offers many valuable studies that can guide us in designing more efficient networks. 
We will leave this part of the research for future exploration.

\section{Conclusion}

Our work demonstrates that neural networks in the pruning-during-training process reflect similar manners as the brain's information propagation mechanisms, and validates the effectiveness of data compression for mitigating the performance decline caused by model pruning. 
We hope our findings will bring insights to the future compression methods and inspire more interaction between research of neuroscience and computer science.

%
%
\bibliographystyle{splncs04}
\bibliography{references}

\begin{thebibliography}{10}
\providecommand{\url}[1]{\texttt{#1}}
\providecommand{\urlprefix}{URL }
\providecommand{\doi}[1]{https://doi.org/#1}

\bibitem{ahmad2019can}
Ahmad, S., Scheinkman, L.: How can we be so dense? the benefits of using highly sparse representations. arXiv preprint arXiv:1903.11257  (2019)

\bibitem{alayrac2022flamingo}
Alayrac, J.B., Donahue, J., Luc, P., Miech, A., Barr, I., Hasson, Y., Lenc, K., Mensch, A., Millican, K., Reynolds, M., et~al.: Flamingo: a visual language model for few-shot learning. Advances in neural information processing systems  \textbf{35},  23716--23736 (2022)

\bibitem{barrett2016optimal}
Barrett, D.G., Deneve, S., Machens, C.K.: Optimal compensation for neuron loss. Elife  \textbf{5},  e12454 (2016)

\bibitem{bolya2022token}
Bolya, D., Fu, C.Y., Dai, X., Zhang, P., Feichtenhofer, C., Hoffman, J.: Token merging: Your vit but faster. arXiv preprint arXiv:2210.09461  (2022)

\bibitem{effectsofvisualworkingmemoryload}
Brockhoff, L., Vetter, L., Bruchmann, M., Schindler, S., Moeck, R., Straube, T.: The effects of visual working memory load on detection and neural processing of task-unrelated auditory stimuli. Scientific Reports  \textbf{13} (03 2023). \doi{10.1038/s41598-023-31132-7}

\bibitem{bu2021population}
Bu, Y., Gao, W., Zou, S., Veeravalli, V.V.: Population risk improvement with model compression: An information-theoretic approach. Entropy  \textbf{23}(10), ~1255 (2021)

\bibitem{chen2021crossvit}
Chen, C.F.R., Fan, Q., Panda, R.: Crossvit: Cross-attention multi-scale vision transformer for image classification. In: Proceedings of the IEEE/CVF international conference on computer vision. pp. 357--366 (2021)

\bibitem{chen2021pre}
Chen, H., Wang, Y., Guo, T., Xu, C., Deng, Y., Liu, Z., Ma, S., Xu, C., Xu, C., Gao, W.: Pre-trained image processing transformer. In: Proceedings of the IEEE/CVF conference on computer vision and pattern recognition. pp. 12299--12310 (2021)

\bibitem{chen2021chasing}
Chen, T., Cheng, Y., Gan, Z., Yuan, L., Zhang, L., Wang, Z.: Chasing sparsity in vision transformers: An end-to-end exploration. Advances in Neural Information Processing Systems  \textbf{34},  19974--19988 (2021)

\bibitem{chen2022dddm}
Chen, X., Li, X., Zhou, Y., Yang, T.: Dddm: a brain-inspired framework for robust classification. arXiv preprint arXiv:2205.10117  (2022)

\bibitem{chen2022fast}
Chen, Y., Gu, X., Liu, Z., Liang, J.: A fast inference vision transformer for automatic pavement image classification and its visual interpretation method. Remote Sensing  \textbf{14}(8), ~1877 (2022)

\bibitem{devlin2018bert}
Devlin, J., Chang, M.W., Lee, K., Toutanova, K.: Bert: Pre-training of deep bidirectional transformers for language understanding. arXiv preprint arXiv:1810.04805  (2018)

\bibitem{Dong_2022_CVPR}
Dong, X., Bao, J., Chen, D., Zhang, W., Yu, N., Yuan, L., Chen, D., Guo, B.: Cswin transformer: A general vision transformer backbone with cross-shaped windows. In: Proceedings of the IEEE/CVF Conference on Computer Vision and Pattern Recognition (CVPR). pp. 12124--12134 (June 2022)

\bibitem{dosovitskiy2021image}
Dosovitskiy, A., Beyer, L., Kolesnikov, A., Weissenborn, D., Zhai, X., Unterthiner, T., Dehghani, M., Minderer, M., Heigold, G., Gelly, S., Uszkoreit, J., Houlsby, N.: An image is worth 16x16 words: Transformers for image recognition at scale (2021)

\bibitem{dziugaite2017computing}
Dziugaite, G.K., Roy, D.M.: Computing nonvacuous generalization bounds for deep (stochastic) neural networks with many more parameters than training data. arXiv preprint arXiv:1703.11008  (2017)

\bibitem{foldiak2003sparse}
Foldiak, P.: Sparse coding in the primate cortex. The handbook of brain theory and neural networks  (2003)

\bibitem{gao2019rate}
Gao, W., Liu, Y.H., Wang, C., Oh, S.: Rate distortion for model compression: From theory to practice. In: International Conference on Machine Learning. pp. 2102--2111. PMLR (2019)

\bibitem{girdhar2023imagebind}
Girdhar, R., El-Nouby, A., Liu, Z., Singh, M., Alwala, K.V., Joulin, A., Misra, I.: Imagebind: One embedding space to bind them all. In: Proceedings of the IEEE/CVF Conference on Computer Vision and Pattern Recognition. pp. 15180--15190 (2023)

\bibitem{hatamizadeh2022unetformer}
Hatamizadeh, A., Xu, Z., Yang, D., Li, W., Roth, H., Xu, D.: Unetformer: A unified vision transformer model and pre-training framework for 3d medical image segmentation. arXiv preprint arXiv:2204.00631  (2022)

\bibitem{doi:10.1073/pnas.1918578117}
Hecker, A., Schulze, W., Oster, J., Richter, D.O., Schuster, S.: Removing a single neuron in a vertebrate brain forever abolishes an essential behavior. Proceedings of the National Academy of Sciences  \textbf{117}(6),  3254--3260 (2020). \doi{10.1073/pnas.1918578117}, \url{https://www.pnas.org/doi/abs/10.1073/pnas.1918578117}

\bibitem{horvath2021manipulation}
Horv{\'a}th, J., Baireddy, S., Hao, H., Montserrat, D.M., Delp, E.J.: Manipulation detection in satellite images using vision transformer. In: Proceedings of the IEEE/CVF conference on computer vision and pattern recognition. pp. 1032--1041 (2021)

\bibitem{hou2022multi}
Hou, Z., Kung, S.Y.: Multi-dimensional model compression of vision transformer. In: 2022 IEEE International Conference on Multimedia and Expo (ICME). pp. 01--06. IEEE (2022)

\bibitem{kirillov2023segany}
Kirillov, A., Mintun, E., Ravi, N., Mao, H., Rolland, C., Gustafson, L., Xiao, T., Whitehead, S., Berg, A.C., Lo, W.Y., Doll{\'a}r, P., Girshick, R.: Segment anything. arXiv:2304.02643  (2023)

\bibitem{kong2022spvit}
Kong, Z., Dong, P., Ma, X., Meng, X., Niu, W., Sun, M., Shen, X., Yuan, G., Ren, B., Tang, H., et~al.: Spvit: Enabling faster vision transformers via latency-aware soft token pruning. In: European conference on computer vision. pp. 620--640. Springer (2022)

\bibitem{Blindedbytheload}
Lavie, N., Beck, D., Konstantinou, N.: Blinded by the load: Attention, awareness and the role of perceptual load. Philosophical transactions of the Royal Society of London. Series B, Biological sciences  \textbf{369},  20130205 (05 2014). \doi{10.1098/rstb.2013.0205}

\bibitem{lee2018snip}
Lee, N., Ajanthan, T., Torr, P.H.: Snip: Single-shot network pruning based on connection sensitivity. arXiv preprint arXiv:1810.02340  (2018)

\bibitem{li2023constraint}
Li, J., Zhang, L.L., Xu, J., Wang, Y., Yan, S., Xia, Y., Yang, Y., Cao, T., Sun, H., Deng, W., et~al.: Constraint-aware and ranking-distilled token pruning for efficient transformer inference. In: Proceedings of the 29th ACM SIGKDD Conference on Knowledge Discovery and Data Mining. pp. 1280--1290 (2023)

\bibitem{Li_2022_CVPR}
Li, Y., Wu, C.Y., Fan, H., Mangalam, K., Xiong, B., Malik, J., Feichtenhofer, C.: Mvitv2: Improved multiscale vision transformers for classification and detection. In: Proceedings of the IEEE/CVF Conference on Computer Vision and Pattern Recognition (CVPR). pp. 4804--4814 (June 2022)

\bibitem{Lin2020Dynamic}
Lin, T., Stich, S.U., Barba, L., Dmitriev, D., Jaggi, M.: Dynamic model pruning with feedback. In: International Conference on Learning Representations (2020), \url{https://openreview.net/forum?id=SJem8lSFwB}

\bibitem{macnamara2012electrocortical}
MacNamara, A., Schmidt, J., Zelinsky, G.J., Hajcak, G.: Electrocortical and ocular indices of attention to fearful and neutral faces presented under high and low working memory load. Biological Psychology  \textbf{91}(3),  349--356 (2012)

\bibitem{mocanu2018scalable}
Mocanu, D.C., Mocanu, E., Stone, P., Nguyen, P.H., Gibescu, M., Liotta, A.: Scalable training of artificial neural networks with adaptive sparse connectivity inspired by network science. Nature communications  \textbf{9}(1), ~2383 (2018)

\bibitem{olshausen1996emergence}
Olshausen, B.A., Field, D.J.: Emergence of simple-cell receptive field properties by learning a sparse code for natural images. Nature  \textbf{381}(6583),  607--609 (1996)

\bibitem{rao2021dynamicvit}
Rao, Y., Zhao, W., Liu, B., Lu, J., Zhou, J., Hsieh, C.J.: Dynamicvit: Efficient vision transformers with dynamic token sparsification. Advances in neural information processing systems  \textbf{34},  13937--13949 (2021)

\bibitem{tanaka2020pruning}
Tanaka, H., Kunin, D., Yamins, D.L., Ganguli, S.: Pruning neural networks without any data by iteratively conserving synaptic flow. Advances in neural information processing systems  \textbf{33},  6377--6389 (2020)

\bibitem{touvron2021training}
Touvron, H., Cord, M., Douze, M., Massa, F., Sablayrolles, A., J{\'e}gou, H.: Training data-efficient image transformers \& distillation through attention. In: International conference on machine learning. pp. 10347--10357. PMLR (2021)

\bibitem{troscianko2023model}
Troscianko, J., Osorio, D.: A model of colour appearance based on efficient coding of natural images. PLOS Computational Biology  \textbf{19}(6),  e1011117 (2023)

\bibitem{vaswani2017attention}
Vaswani, A., Shazeer, N., Parmar, N., Uszkoreit, J., Jones, L., Gomez, A.N., Kaiser, {\L}., Polosukhin, I.: Attention is all you need. Advances in neural information processing systems  \textbf{30} (2017)

\bibitem{wang2023cait}
Wang, A., Chen, H., Lin, Z., Zhao, S., Han, J., Ding, G.: Cait: Triple-win compression towards high accuracy, fast inference, and favorable transferability for vits. arXiv preprint arXiv:2309.15755  (2023)

\bibitem{wang2022bridged}
Wang, Y., Ye, T., Cao, L., Huang, W., Sun, F., He, F., Tao, D.: Bridged transformer for vision and point cloud 3d object detection. In: Proceedings of the IEEE/CVF conference on computer vision and pattern recognition. pp. 12114--12123 (2022)

\bibitem{wang2022vtc}
Wang, Z., Luo, H., Wang, P., Ding, F., Wang, F., Li, H.: Vtc-lfc: Vision transformer compression with low-frequency components. Advances in Neural Information Processing Systems  \textbf{35},  13974--13988 (2022)

\bibitem{xu2022evo}
Xu, Y., Zhang, Z., Zhang, M., Sheng, K., Li, K., Dong, W., Zhang, L., Xu, C., Sun, X.: Evo-vit: Slow-fast token evolution for dynamic vision transformer. In: Proceedings of the AAAI Conference on Artificial Intelligence. vol.~36, pp. 2964--2972 (2022)

\bibitem{yang2020learning}
Yang, F., Yang, H., Fu, J., Lu, H., Guo, B.: Learning texture transformer network for image super-resolution (2020)

\bibitem{yang2022lavt}
Yang, Z., Wang, J., Tang, Y., Chen, K., Zhao, H., Torr, P.H.: Lavt: Language-aware vision transformer for referring image segmentation. In: Proceedings of the IEEE/CVF Conference on Computer Vision and Pattern Recognition. pp. 18155--18165 (2022)

\bibitem{yu2022unified}
Yu, S., Chen, T., Shen, J., Yuan, H., Tan, J., Yang, S., Liu, J., Wang, Z.: Unified visual transformer compression (2022)

\bibitem{yuan2022volo}
Yuan, L., Hou, Q., Jiang, Z., Feng, J., Yan, S.: Volo: Vision outlooker for visual recognition. IEEE transactions on pattern analysis and machine intelligence  \textbf{45}(5),  6575--6586 (2022)

\bibitem{zheng2022savit}
Zheng, C., Zhang, K., Yang, Z., Tan, W., Xiao, J., Ren, Y., Pu, S., et~al.: Savit: Structure-aware vision transformer pruning via collaborative optimization. Advances in Neural Information Processing Systems  \textbf{35},  9010--9023 (2022)

\bibitem{zhou2018non}
Zhou, W., Veitch, V., Austern, M., Adams, R.P., Orbanz, P.: Non-vacuous generalization bounds at the imagenet scale: a pac-bayesian compression approach. arXiv preprint arXiv:1804.05862  (2018)

\bibitem{zhu2021vision}
Zhu, M., Tang, Y., Han, K.: Vision transformer pruning (2021)

\end{thebibliography}

\end{document}